\documentclass[pdflatex,sn-basic]{sn-jnl}

\usepackage{graphicx}%
\usepackage{multirow}%
\usepackage{amsmath,amssymb,amsfonts}%
\usepackage{amsthm}%
\usepackage{mathrsfs}%
\usepackage[title]{appendix}%
\usepackage{xcolor}%
\usepackage{textcomp}%
\usepackage{manyfoot}%
\usepackage{booktabs}%
\usepackage{algorithm}%
\usepackage{algorithmicx}%
\usepackage{algpseudocode}%
\usepackage{listings}%

\theoremstyle{thmstyleone}%
\theoremstyle{thmstyletwo}%

\theoremstyle{thmstylethree}%

\begin{document}

\title[Article Title]{VDLF‑Net:  Variational Feature Fusion for Adaptive and Few‑Shot Visual  Learning}


\author*[1]{\fnm{First} \sur{Jiawei Yan}}\email{yjw1998@sjtu.edu.cn}



\affil*[1]{\orgdiv{Department of Statistics}, \orgname{Shanghai Jiao Tong University}, \orgaddress{ \city{Shanghai}, \postcode{200240}, \country{China}}}




\abstract{This paper introduces VDLF-Net, which attaches a compact VAE to a multi-scale CNN backbone. Latent vectors and softmax-gate support the backbone feature maps, while $\ell_2$-normalized embeddings from the gated maps contribute toward supervised classification or episodic few-shot prediction. Under standard CIFAR-100 and Mini-ImageNet protocols, VDLF-Net demonstrates an improved performance over ResNet-50 Enhanced, VGG-16, Prototypical Networks, and Matching Networks. Extensive ablations show that removing the fine-resolution scale has the greatest impact on VDLF-Net's performance. At the same time, KL and reconstruction at the chosen $\alpha$ pose a minor performance reduction, demonstrating that performance gains over classical episodic baselines mainly originate from the full VDLF-Net architecture and training strategy.}




\maketitle

\section{Introduction}\label{sec1}

This paper studies one network applicable in two standard settings: supervised CIFAR-100 classification and Mini-ImageNet few-shot episodes. The model performs input-conditioned fusion of multi-scale convolutional features through a small VAE bottleneck and shared heads for both settings.

High-dimensional visual problems, from medical imaging to remote sensing, demand adaptive feature design and data-efficient adaptation \citep{Litjens_2017,2021A,2017Squeeze,2020Generalizing,2024A}. Therefore, we benchmark on CIFAR-100 and Mini-ImageNet, where splits, budgets, and baselines are standardized, ensuring fair comparisons and claims that attach to these recognition protocols rather than specialized deployments.

Despite the extensive usage of machine learning methods, these suffer from various limitations. For instance, CNNs learn strong features but underuse uncertainty in low-data regimes \citep{kendall2017uncertainties}. VAEs regularize latent structure but are often bolted on with fixed merges \citep{2014Auto}. In contrast, Few-shot methods (metric \citep{snell2017prototypicalnetworksfewshotlearning}, and optimization-based \citep{finn2017modelagnosticmetalearningfastadaptation}) typically fuse scales by hand or freeze encoders, and adaptive fusion and episodic objectives often remain loosely coupled \citep{Schmidhuber_2015}.

\textbf{VDLF-Net (one-sentence sketch).} Driven by the limitations of the existing solutions, in the proposed VDLF-Net, the average-pooled multi-scale features feed a VAE, latent draws softmax-gate the same scale tensors, and the $\ell_2$-normalized vectors enter either a classifier or episodic prototypes. The VDLF-Net is trained with task loss, reconstruction loss, and KL divergence.

\noindent Contributions of the developed VDLF-Net are: (1)~unified variational--deep fusion for supervised and episodic training; (2)~FAAM: latent-gated multi-scale pooling; (3)~few-shot training with Monte-Carlo support draws and cosine prototypes; (4)~benchmarks and ablations (Section~\ref{sec: Experimental Design}, Subsection~\ref{subsec:ablation}).

The remainder of this paper is organized as follows. Section~\ref{sec: Variational-Deep Learning Fusion Framework} presents VDLF-Net: multi-scale CNN--VAE coupling, the Feature-Adaptive Approximation Mechanism (FAAM), and variationally regularized few-shot inference. Section~\ref{sec: Experimental Design} describes the datasets employed, evaluation protocols, baselines, and training details. Section~\ref{sec: Results and Analysis} reports main comparisons and component ablation studies. Section~\ref{sec: Conclusion} summarizes this study and discusses the limitations.

\subsection{Related work and positioning}
\label{subsec:related_work}

\noindent\textbf{Related work.} CNNs, attention, and dynamic layers \citep{CiresanMMGS11, HeZRS16, 2017Squeeze, JiaBTG16} are usually trained discriminatively. 
On the other hand, VAEs \citep{2014Auto} add a latent structure but rarely drive scale-wise fusion in vision. Besides, Prototypical and Matching nets \citep{snell2017prototypicalnetworksfewshotlearning,vinyals2016matching}, relation modules \citep{sung2018learning}, and MAML-style updates \citep{finn2017modelagnosticmetalearningfastadaptation} set few-shot practice, and \citep{chen2019closerlook,triantafillou2020meta} emphasise fair evaluation. VDLF-Net stays in the prototype family but lets latent variables set multi-scale weights instead of a fixed merge.

\section{Variational-Deep Learning Fusion Framework}
\label{sec: Variational-Deep Learning Fusion Framework}

\subsection{Intuition and modeling goal}
\label{subsec:intuition}

This study seeks a single representation in which (i) multi-scale CNN evidence is combined in a way that depends on the input, not only on a fixed merge rule, and (ii) uncertainty about that merge is represented so that episodic decisions can average over plausible fusions. Averaging features alone answers (i) only weakly, while a standalone VAE on a frozen merge answers (ii) but not (i). Therefore, VDLF-Net runs a lightweight VAE on an initial fused summary $\mathbf{F}_{\text{fused}}^{(0)}$, draws latent codes $\mathbf{z}$, and uses each draw to form nonnegative weights over the same scale-wise tensors that produced the summary. Reconstruction and KL terms keep the latent process identifiable, while the task loss on $\mathbf{F}_{\text{norm}}$ ties the uncertainty to discriminative structure.

\subsection{Overall Framework Design}
\label{subsec: Overall Framework Design}

\begin{figure}[htp]
    \centering
    \includegraphics[width=0.8\textwidth]{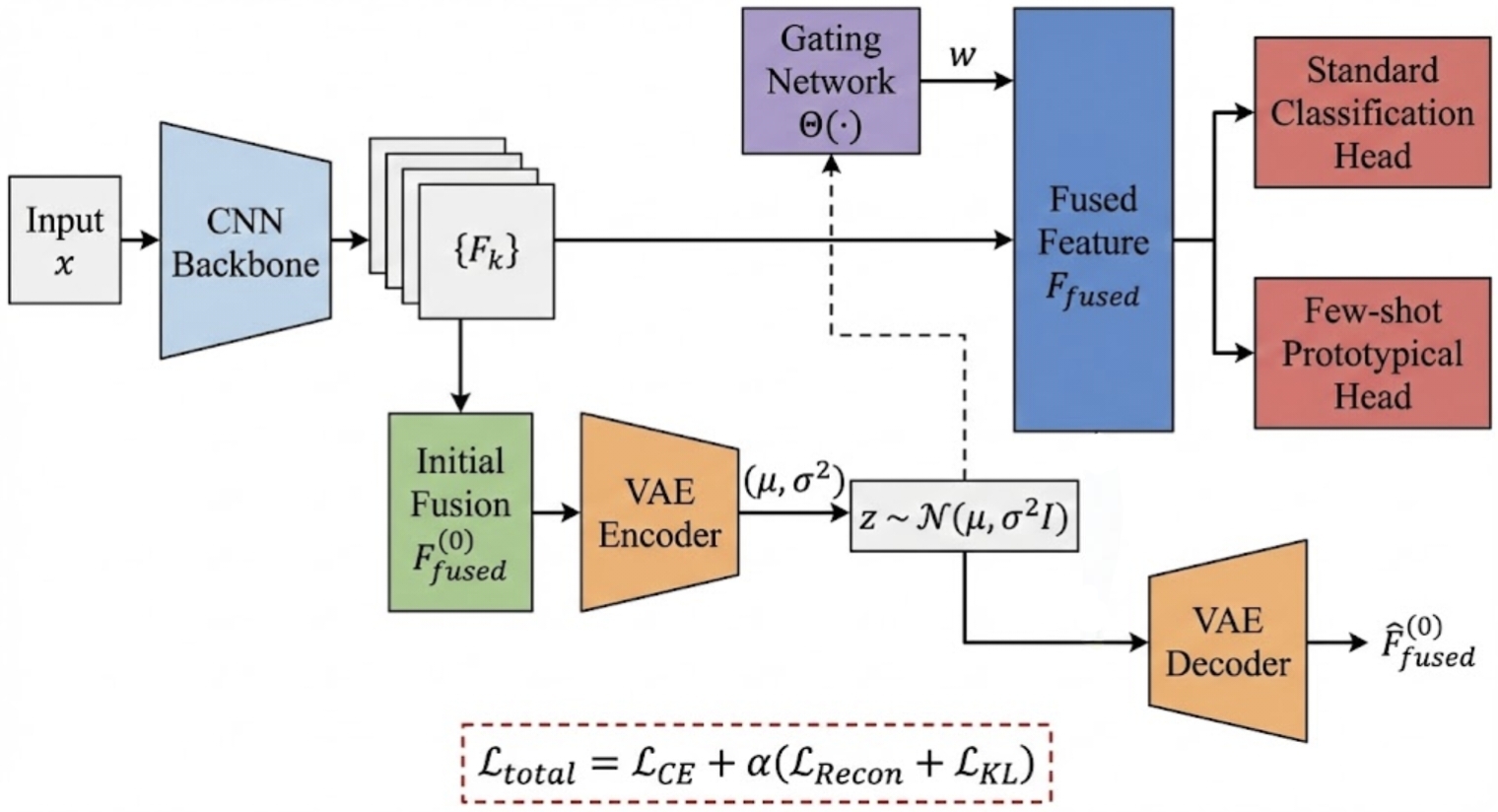}
    \caption{Overview of the Variational-Deep Learning Fusion Network (VDLF-Net).}
\label{fig:overall_design}
\end{figure}

Fig. \ref{fig:overall_design} illustrates VDLF-Net, which comprises the following components:

\begin{enumerate}
\item \textbf{Deep Representation Stream.} Given an input image $\mathbf{x}$, a CNN backbone extracts multi-scale feature maps $\{\mathbf{F}_k\}_{k=1}^{K}$ from selected convolutional layers of the backbone, capturing various spatial resolutions and semantic information \citep{CiresanMMGS11, HeZRS16}. These multi-scale features serve as inputs to the fusion module, which produces a task-adaptive fused representation.

\item \textbf{Variational Inference Stream.} The initial fused feature $\mathbf{F}_{\text{fused}}^{(0)}$ is calculated using a fixed initialization rule, e.g., simple averaging, which is then passed to a VAE encoder. The encoder parameterizes an approximate posterior over the fused feature and samples a latent code $\mathbf{z}$ via the reparameterization process \citep{2014Auto}, providing uncertainty-aware latent information for adaptive feature fusion:
\begin{align}
   q_\phi(\mathbf{z} \mid \mathbf{F}_{\text{fused}}^{(0)}) = \mathcal{N}(\boldsymbol{\mu}, \operatorname{diag}(\boldsymbol{\sigma}^2)). 
\end{align}

\item \textbf{Bidirectional Fusion.} Bottom-up, $\mathbf{F}_{\text{fused}}^{(0)}$ is encoded to $\mathbf{z}$; top-down, $\mathbf{z}$ drives $\Theta(\cdot)$ to produce softmax weights $\mathbf{w}$ over the same $\{\mathbf{F}_k\}$, yielding $\mathbf{F}_{\text{fused}}$ and normalized $\mathbf{F}_{\text{norm}}$. Both directions are needed, as the encoder summarises cross-scale context for the gate, and the gate maps latent uncertainty into which scale matters.

\item \textbf{Task heads.}
The model supports two training/evaluation protocols via corresponding task heads:
\begin{itemize}
\item \textbf{Standard classification head} for conventional supervised learning, optimized by a supervised cross-entropy loss $\mathcal{L}_{\text{CE}}^{\text{sup}}$.
\item \textbf{Few-shot prototypical head} for episodic inference (Subsection~\ref{subsec: Variationally-Regularized Few-Shot Inference}), optimized by an episodic cross-entropy loss $\mathcal{L}_{\text{CE}}^{\text{epi}}$ defined on query samples.
\end{itemize}

\item \textbf{Unified Optimization Objective.}
VDLF-Net is optimized end-to-end using a unified objective that couples a task-specific discriminative loss with variational regularization:
\begin{align}
\mathcal{L}_{\text{total}} = \mathcal{L}_{\text{task}} + \alpha \left(\mathcal{L}_{\text{Recon}} + \mathcal{L}_{\text{KL}}\right),
\end{align}
where $\mathcal{L}_{\text{task}} \in \{\mathcal{L}_{\text{CE}}^{\text{sup}}, \mathcal{L}_{\text{CE}}^{\text{epi}}\}$ depends on whether training follows the standard supervised protocol or the episodic few-shot protocol. Parameter $\alpha$ is a balance coefficient that controls the relative importance of the variational regularization terms (reconstruction and KL divergence) in the total loss. Further to that, the reconstruction term $\mathcal{L}_{\text{Recon}}$ is computed by a VAE decoder $g_\psi(\cdot)$, which reconstructs the initial fused feature from the latent vector $\mathbf{z}$:
\begin{align}
\hat{\mathbf{F}}_{\text{fused}}^{(0)} = g_\psi(\mathbf{z}), \quad
\mathcal{L}_{\text{Recon}} = \left\| \mathbf{F}_{\text{fused}}^{(0)} - \hat{\mathbf{F}}_{\text{fused}}^{(0)} \right\|_2^2.
\end{align}
The KL divergence term $\mathcal{L}_{\text{KL}}$ regularizes the posterior to match the prior:
\begin{align}
\mathcal{L}_{\text{KL}} = D_{\text{KL}}\left(q_\phi(\mathbf{z} \mid \mathbf{F}_{\text{fused}}^{(0)}) \| p(\mathbf{z}) \right), \quad
p(\mathbf{z}) = \mathcal{N}(\mathbf{0}, \mathbf{I}),
\end{align}
encouraging a well-structured latent space and improving generalization in both supervised and few-shot regimes.
\end{enumerate}

\textbf{Gating network instantiation.} $\Theta$ is a two-layer MLP, i.e., a fully connected map $\mathbf{z}\rightarrow \mathrm{ReLU}(W_1\mathbf{z}+\mathbf{b}_1)$ followed by $W_2(\cdot)+\mathbf{b}_2$ and softmax over $K$ logits. The CIFAR-100 model uses a hidden width of $128$, and the Mini-ImageNet episodic implementation uses the same depth with hidden width $64$ when operating on $84\times 84$ inputs.

\subsection{Feature-Adaptive Approximation Mechanism}
\label{subsec: Feature-Adaptive Approximation Mechanism}

FAAM instantiates Fig.~\ref{fig:overall_design} for episodic training: latent draws $\mathbf{z}^{(t)}$ produce fusion weights, then centered $\ell_2$ normalization yields unit-norm features for cosine matching.

\begin{figure}[htp]
    \centering
\includegraphics[width=0.8\textwidth]{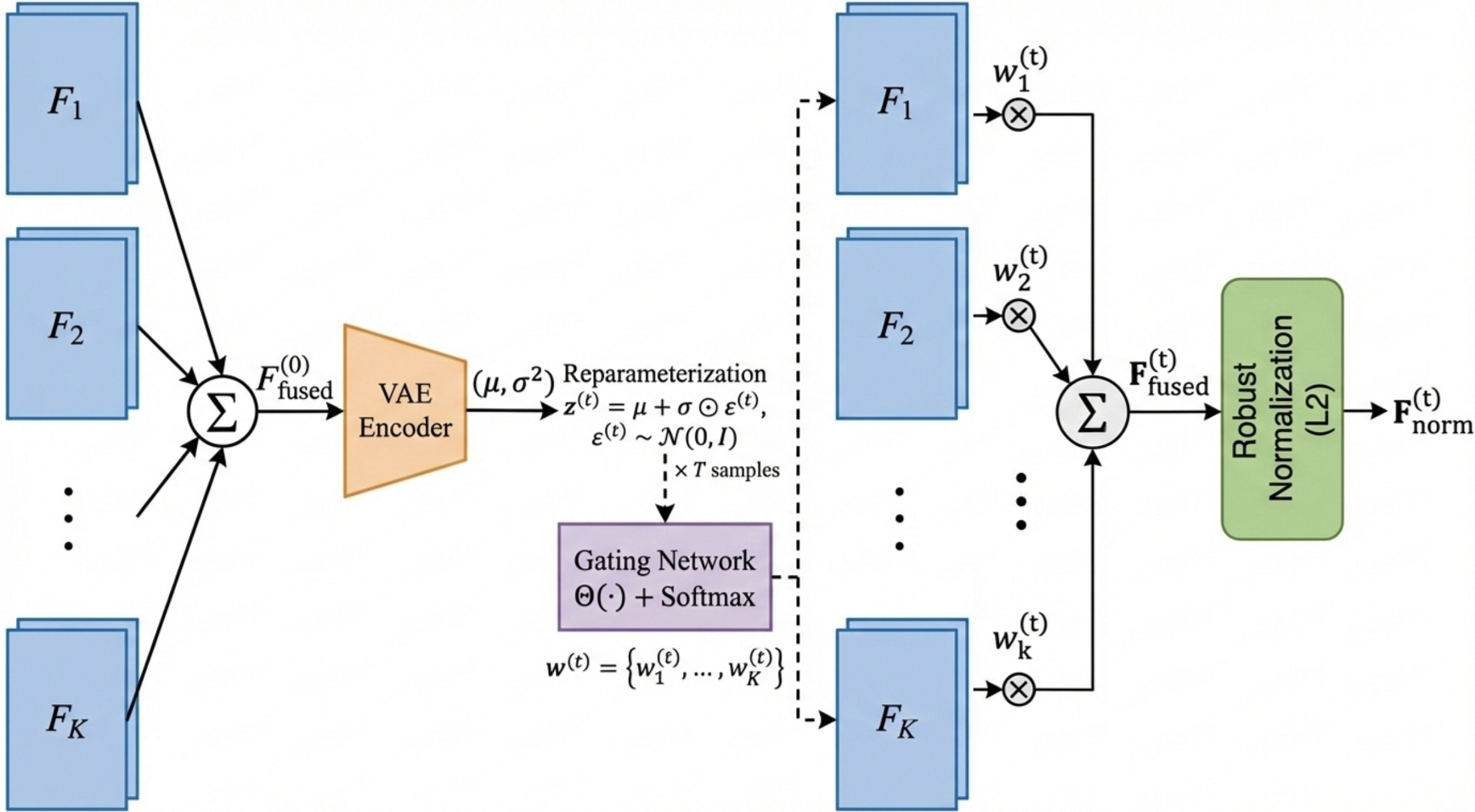}
    \caption{Feature-Adaptive Approximation Mechanism (FAAM).}
    \label{fig:faam}
\end{figure}

For each support image, sample $T$ has the following latent samples:
\begin{align}
\mathbf{z}^{(t)}=\boldsymbol{\mu}+\boldsymbol{\sigma}\odot \boldsymbol{\epsilon}^{(t)},\quad
\boldsymbol{\epsilon}^{(t)}\sim\mathcal{N}(\mathbf{0},\mathbf{I}),\quad t=1,\ldots,T.
\end{align}
These latent samples condition the gating network $\Theta$, which generates $T$ distinct fusion weights, $\mathbf{w}^{(t)}$, typically via a Softmax function:
\begin{align}
\mathbf{w}^{(t)}=\text{Softmax}(\Theta(\mathbf{z}^{(t)})),\quad \mathbf{w}^{(t)}\in\mathbb{R}^{K}.
\end{align}
The fused feature for each sample is computed as:
\begin{align}
\mathbf{F}_{\text{fused}}^{(t)}=\sum_{k=1}^{K} w_k^{(t)}\cdot \mathbf{F}_k,
\end{align}
followed by centering and $L_2$ normalization:
\begin{align}
\mathbf{F}_{\text{norm}}^{(t)}=
\frac{\mathbf{F}_{\text{fused}}^{(t)}-\mu_{\text{fused}}}
{\left\|\mathbf{F}_{\text{fused}}^{(t)}-\mu_{\text{fused}}\right\|_2+\epsilon},
\end{align}
where $\mu_{\text{fused}}$ is the batch mean and $\epsilon>0$ ensures numerical stability.

\subsection{Variationally-Regularized Few-Shot Inference}
\label{subsec: Variationally-Regularized Few-Shot Inference}

Queries use a single deterministic forward pass, supporting an average of $T$ FAAM samples per image when forming prototypes.

Each episode $\mathcal{E}$ is an $N$-way $K$-shot task with support
$\mathcal{S}=\{(\mathbf{x}_{c,i}, y_{c,i}=c)\}_{c=1,i=1}^{N,K}$
and a query set $\mathcal{Q}=\{(\mathbf{x}_q,y_q)\}$.

We also generate $T$ latent samples for each support example to improve robustness under label scarcity, yielding
$T$ embeddings $\mathbf{F}_{\text{norm}}^{(t)}(\mathbf{x}_{c,i})$ that share the same class label $c$.
The class prototype is computed as
\begin{align}
   \mathbf{P}_c=\frac{1}{KT}\sum_{i=1}^{K}\sum_{t=1}^{T}\mathbf{F}_{\text{norm}}^{(t)}(\mathbf{x}_{c,i}),
\quad c\in\{1,\ldots,N\}. 
\end{align}

For each query, a deterministic embedding is used, denoted as $\mathbf{F}_{\text{norm}}(\mathbf{x}_q)$. The cosine similarity is formulated as
\begin{align}
\operatorname{sim}(\mathbf{a},\mathbf{b})
=
\frac{\mathbf{a}^{\top}\mathbf{b}}{\|\mathbf{a}\|_2\|\mathbf{b}\|_2},
\end{align}
and prediction is performed using nearest-prototype matching:
\begin{align}
\hat{y}_q=\arg\max_{c\in\{1,\ldots,N\}}
\operatorname{sim}\!\left(\mathbf{F}_{\text{norm}}(\mathbf{x}_q),\mathbf{P}_c\right).
\end{align}

During episodic meta-training, the episodic task loss is defined as $\mathcal{L}_{\text{task}} \triangleq \mathcal{L}_{\text{CE}}^{\text{epi}}$ on the query set using temperature-scaled cosine-softmax logits:
\begin{align}
p(y=c\mid \mathbf{x}_q,\mathcal{S})
=
\frac{\exp\!\left(\tau\cdot \operatorname{sim}\!\left(\mathbf{F}_{\text{norm}}(\mathbf{x}_q),\mathbf{P}_c\right)\right)}
{\sum_{c'=1}^{N}\exp\!\left(\tau\cdot \operatorname{sim}\!\left(\mathbf{F}_{\text{norm}}(\mathbf{x}_q),\mathbf{P}_{c'}\right)\right)},
\end{align}
\begin{align}
\mathcal{L}_{\text{CE}}^{\text{epi}}(\mathcal{E})
=
-\frac{1}{|\mathcal{Q}|}\sum_{(\mathbf{x}_q,y_q)\in\mathcal{Q}}
\log p(y=y_q\mid \mathbf{x}_q,\mathcal{S}),
\end{align}
where $\tau>0$ is a fixed temperature parameter.

Finally, consistent with the unified objective in Subsection~\ref{subsec: Overall Framework Design}, the episode-level total loss is
\begin{align}
\mathcal{L}_{\text{total}}(\mathcal{E})
=
\mathcal{L}_{\text{CE}}^{\text{epi}}(\mathcal{E})
+
\alpha\Big(\mathcal{L}_{\text{Recon}}(\mathcal{E})+\mathcal{L}_{\text{KL}}(\mathcal{E})\Big),
\end{align}
and training minimizes $\mathbb{E}_{\mathcal{E}}\big[\mathcal{L}_{\text{total}}(\mathcal{E})\big]$, where $\mathcal{L}_{\text{CE}}^{\text{epi}}(\mathcal{E})$ is averaged over query samples. In contrast, $\mathcal{L}_{\text{Recon}}(\mathcal{E})$ and $\mathcal{L}_{\text{KL}}(\mathcal{E})$ are computed and averaged over all samples within the episode (support and query).

\subsection{Computational cost}
\label{subsec:complexity}

Let $B$ denote batch size (supervised) or the number of images processed in an inner loop, and $C_{\text{back}}$ be the cost of the truncated ResNet-50 backbone up to multi-scale pooling. One forward pass applies $C_{\text{back}}$, pools to $K$ tensors, averages to form $\mathbf{F}_{\text{fused}}^{(0)}$, and runs the VAE encoder/decoder and gating MLP with $\mathcal{O}(d\,d_h + d_h K)$ parameters, where $d$ is the fused feature dimension and $d_h$ is the gate hidden width. This is negligible relative to $C_{\text{back}}$ for typical $K{\le}4$. In episodic evaluation with $T$ latent draws per support image, support processing scales linearly in $T$ relative to a single draw, whereas query embeddings use one deterministic forward without repeated sampling. Notably, memory is dominated by activations of the backbone and by storing $K$ scale-wise tensors until fusion.

\section{Experimental Design}
\label{sec: Experimental Design}

\subsection{Datasets and Settings}
\label{subsec: Datasets and Settings}

\textbf{Datasets:}
Two widely used benchmarks spanning standard classification and few-shot learning are used:
\begin{itemize}
    \item \textbf{CIFAR-100:} 60{,}000 color images of size $32 \times 32$ evenly distributed over 100 classes. The standard train/test split for supervised evaluation is adopted.
    
    \item \textbf{Mini-ImageNet:} A standard few-shot benchmark constructed from ImageNet, containing 100 classes with 600 images per class. Following the conventional protocol, we use 64 classes for episodic training, 16 for validation, and 20 for testing (unseen during training), consistent with Subsection~\ref{subsec: Variationally-Regularized Few-Shot Inference}.
\end{itemize}

\textbf{Task Settings:}
\begin{itemize}
    \item \textbf{Standard Classification:} Models are trained and evaluated on the full train/test splits of CIFAR-100 under a supervised protocol.
    
    \item \textbf{Few-Shot Inference:} The standard episodic protocol is adopted. Each episode is an $N$-way $K$-shot task with a support set of size $N \times K$ and a query set of size $N \times Q$. We set $N=5$, $K \in \{1,5\}$, and $Q=15$ queries per class (Table~\ref{tab:hyperparams}), consistent with common Mini-ImageNet configurations in public implementations.
\end{itemize}

\begin{table}[htbp]
\centering
\caption{Hyperparameters and protocol summary for the reported results (Tables~\ref{tab:standard_results}--\ref{tab:fewshot_results}) and for Mini-ImageNet ablations (Table~\ref{tab:ablation_min}).}
\label{tab:hyperparams}
\small
\setlength{\tabcolsep}{4pt}
\begin{tabular}{@{}p{0.38\textwidth}p{0.52\textwidth}@{}}
\toprule
\textbf{Item} & \textbf{Value} \\
\midrule
\multicolumn{2}{@{}l}{\textit{CIFAR-100 supervised}} \\
Optimizer / schedule & AdamW; cosine decay with $\eta_{\min}=0.1\times$ initial lr \\
Learning rate / batch / epochs & $0.002$ / $128$ / $180$ \\
Weight decay / $\alpha$ & $1.5\times 10^{-4}$ / $0.01$ \\
Backbone / $K$ / latent dim & ResNet-50 (truncated) / $2$ / $128$ \\
Gating MLP hidden width & $128$ \\
\midrule
\multicolumn{2}{@{}l}{\textit{Mini-ImageNet episodic}} \\
$N$-way / $K$-shot / $Q$ & $5$ / $\{1,5\}$ / $15$ \\
Training / test episodes & $150$ / $100$ \\
$T$ / $\tau$ / $\alpha$ & $15$ / $15$ / $0.01$ \\
Optimizer (lr, wd) & AdamW ($10^{-4}$, $5\times 10^{-5}$) \\
Effective batch & $4$ (gradient accumulation) \\
Gating MLP hidden width & $64$ \\
$95\%$ CI & Normal approx.\ on 100 test episodes \\
\midrule
\multicolumn{2}{@{}l}{\textit{Hardware}} \\
GPU & NVIDIA CUDA devices ($\geq 12$\,GB) \\
\bottomrule
\end{tabular}
\end{table}

\textbf{Preprocessing.} Channel normalization, supervised training using resize/crop/flip, and episodic training adding light rotation. For the evaluation, we crop the center of the image.

\subsection{Baseline Methods}
\label{subsec: Baseline Methods}

\begin{itemize}
    \item \textbf{Supervised:} ResNet-50 Enhanced (head matched to our setup) and VGG-16.
    \item \textbf{Few-shot:} Prototypical \citep{snell2017prototypicalnetworksfewshotlearning} and Matching \citep{vinyals2016matching} on Mini-ImageNet under Table~\ref{tab:hyperparams}.
\end{itemize}

\textbf{Scope.} Few-shot rows use Prototypical and Matching nets under the protocol presented in Table~\ref{tab:hyperparams}, which in future work will be extended to newer meta-learners with matched budgets (Section~\ref{subsec:limitations}).

\subsection{Evaluation Metrics}
\label{subsec: Evaluation Metrics}

CIFAR-100: accuracy and macro precision/recall/F1. Few-shot: episode mean accuracy with normal-approximation 95\% intervals (100 test episodes).

\subsection{Implementation Details}
\label{subsec: Implementation Details}

\textbf{Architecture:} VDLF-Net is implemented in PyTorch, using a truncated ResNet-50 backbone before global average pooling. Multi-scale features ($K=2$) 
are extracted from the final convolutional layer (layer4) using two adaptive average pooling operations with spatial resolutions of 2×2 and 1×1, yielding two representations at different scales. The VAE latent dimension is set to 128, and the fusion module follows the latent-conditioned, gating-based aggregation 
described in Section~\ref{sec: Variational-Deep Learning Fusion Framework}, 
producing fused and normalized embeddings $\mathbf{F}_{\text{fused}}$ and $\mathbf{F}_{\text{norm}}$.

\textbf{Training (supervised).} For standard classification on CIFAR-100, 
we optimize $\mathcal{L}_{\text{total}}$ using AdamW with a learning rate of 
0.002, a batch size of 128, and a weight decay of 0.00015 for 180 epochs. 
The loss-balance coefficient $\alpha$ is set to 0.01, selected via empirical 
tuning. CosineAnnealingLR scheduler is also used with $\eta_{\min} = 0.1 \times \text{initial LR}$ 
to ensure smooth learning rate decay throughout training. VGG-16 and ResNet-50 Enhanced 
are trained under the same data preprocessing, augmentation, and evaluation protocol 
with comparable optimization settings.

\textbf{Few-Shot Inference:} For episodic few-shot learning, each episode uses $Q=15$ queries per class. We rely on meta-train with 150 sampled episodes per configuration and evaluate on 100 fresh test episodes. Reported accuracies are episode means; $95\%$ intervals use $\pm 1.96\,\hat{\sigma}/\sqrt{100}$, where $\hat{\sigma}$ is the empirical standard deviation of per-episode accuracies (normal approximation). For support embeddings, FAAM uses $T=15$ Monte-Carlo draws and cosine-softmax temperature $\tau=15$ (Subsection~\ref{subsec: Variationally-Regularized Few-Shot Inference}). The variational weight in the episodic objective is $\alpha{=}0.01$, matching supervised CIFAR-100 and the configuration used to train the Mini-ImageNet model reported in Table~\ref{tab:fewshot_results}. Optimization uses AdamW with learning rate $1 \times 10^{-4}$, weight decay $5 \times 10^{-5}$, and effective batch size 4 via gradient accumulation.

\textbf{Hyperparameter rationale (few-shot).} We use 150/100 train/test episodes and $Q{=}15$ following common Mini-ImageNet practice \citep{snell2017prototypicalnetworksfewshotlearning, vinyals2016matching}. $\tau{=}15$, $T{=}15$, and $\alpha{=}0.01$ were chosen by a small validation grid ($\tau$ sets softmax sharpness; $T$ trades prototype variance for compute, Section~\ref{subsec:complexity}) and to match supervised $\alpha$.

\textbf{Reproducibility and computation.} Seeds were fixed to 42 where possible; 100 test episodes per configuration mitigate variability across runs. Hardware: CUDA GPUs with $\geq 12$\,GB; Table~\ref{tab:hyperparams} records training budgets.

\section{Results and Analysis}
\label{sec: Results and Analysis}

This section reports the supervised CIFAR-100 and Mini-ImageNet few-shot baselines, along with internal ablations.

\subsection{Performance on Standard Classification Tasks}
\label{subsec: Performance on Standard Classification Tasks}

Table~\ref{tab:standard_results} reports CIFAR-100.

\begin{table}[t]
\centering
\caption{Classification performance on CIFAR-100. The best results are highlighted in \textbf{bold}. Precision, recall, and F1 score are all computed using macro averaging.}
\label{tab:standard_results}
\begin{tabular}{lccccc}
\toprule
\textbf{Model} 
& \textbf{Accuracy (\%)} 
& \textbf{Precision (\%)} 
& \textbf{Recall (\%)} 
& \textbf{F1 (\%)} \\
\midrule
VGG-16      & 49.09 & 50.05 & 49.09 & 48.98 \\
ResNet-50 Enhanced   & 56.16 & 56.39 & 56.16 & 55.93 \\
\textbf{VDLF-Net}  & \textbf{58.34} & \textbf{58.72} & \textbf{58.34} & \textbf{58.30} \\
\bottomrule
\end{tabular}
\end{table}

VDLF-Net improves over both CNN baselines on all four metrics; margins vs.\ ResNet-50 enhanced isolate fusion because the head is matched.

\subsection{Few-Shot Inference Performance}
\label{subsec: Few-Shot Inference Performance}

Table~\ref{tab:fewshot_results}: Mini-ImageNet $N$-way $K$-shot (100 test episodes per setting).

\begin{table}[t]
\centering
\setlength{\tabcolsep}{10pt}
\caption{Few-shot classification accuracy (\%; mean $\pm$ 95\% confidence interval) on Mini-ImageNet. The best results are highlighted in \textbf{bold}.}
\label{tab:fewshot_results}
\begin{tabular}{lcc}
\toprule
\textbf{Model} & \textbf{1-shot (\%)} & \textbf{5-shot (\%)} \\
\midrule
Prototypical Networks & $56.80 \pm 1.84$ & $80.09 \pm 1.14$ \\
Matching Networks & $61.87 \pm 2.02$ & $68.87 \pm 1.87$ \\
\textbf{VDLF-Net} & \textbf{70.72$\pm$ 2.03} & \textbf{86.33$\pm$1.17} \\
\bottomrule
\end{tabular}
\end{table}

\begin{table}[htbp]
\centering
\setlength{\tabcolsep}{20pt}
\caption{Mini-ImageNet $5$-way $5$-shot ablation: accuracy (\%; mean $\pm$ approximate $95\%$ CI). ``Full'' uses learned gates and both KL and reconstruction in $\mathcal{L}_{\text{total}}$. Other rows change one structural or loss element at a time.}
\label{tab:ablation_min}
\begin{tabular}{lc}
\toprule
\textbf{Variant} & \textbf{5-shot (\%)} \\
\midrule
Full model & $86.12 \pm 1.15$ \\
w/o KL term & $86.32 \pm 1.14$ \\
w/o reconstruction & $86.08 \pm 1.20$ \\
w/o KL \& reconstruction & $86.27 \pm 1.16$ \\
Uniform scale weights ($1/K$) & $86.01 \pm 1.13$ \\
Fine scale only ($1{\times}1$ branch) & $87.36 \pm 1.13$ \\
Coarse scale only ($2{\times}2$ branch) & $85.28 \pm 1.19$ \\
\bottomrule
\end{tabular}
\end{table}

VDLF-Net leads in 1- and 5-shot, with gains being largest against Matching at 5-shot. Table~\ref{tab:ablation_min} reveals the internal factors leading to these results.

\subsection{Component ablation (Mini-ImageNet 5-way 5-shot)}
\label{subsec:ablation}

Table~\ref{tab:ablation_min} lists single-factor removals trained from scratch under the Table~\ref{tab:hyperparams} budget (150/100 episodes, $T{=}15$, $\tau{=}15$, $\alpha{=}0.01$). The full row (86.12\%) matches Table~\ref{tab:fewshot_results} within sampling noise.

\textbf{Takeaways.} (i) Coarse-only pooling lags markedly, so fine-resolution evidence drives the fusion gain over Table~\ref{tab:fewshot_results}. (ii) Dropping KL or reconstruction barely moves accuracy at $\alpha{=}0.01$; episodic CE dominates. (iii) Learned gates edge uniform mixing (86.12 vs.\ 86.01), implying the lift vs.\ Proto/Matching chiefly comes from the overall stack, not a single loss term.

The fine-only branch can marginally beat the full model ($87.36\%$ vs.\ $86.12\%$): with few scales and 150 meta-episodes, a fixed high-res blend is easier to optimize than a gate yet to peak. This is reported for transparency; it does not erase the coarse-only failure mode nor the headline advantage over external baselines—rather, it flags where more training or regularisation could further justify adaptive fusion.

\subsection{Limitations}
\label{subsec:limitations}

Evidence is limited to CIFAR-100 and Mini-ImageNet, as domain shift and specialized modalities are out of scope. Qualitatively, Table~\ref{tab:ablation_min} shows coarse-only fusion fails first, so fine-detail confusions remain a risk when coarse cues dominate. Furthermore, metric methods degrade when the train/test distributions differ unless the encoder is adapted.

Recent few-shot methods exceed our two episodic baselines in benchmark leaderboards. Thus, we emphasize the use of matched protocols in Table~\ref{tab:hyperparams} and highlight the fusion module as the contribution. Stronger comparisons (additional meta-learners, multi-seed means) are the main follow-up.

Confidence intervals reflect 100 test episodes, highlighting that a single meta-train seed leaves small gaps among KL/Recon rows. Support-side $T$ draws add linear inference cost for wide backbones or detection.

\section{Conclusion}
\label{sec: Conclusion}

This paper addresses the dual challenges of feature-adaptive approximation and few-shot inference for high-dimensional visual data. We introduce VDLF-Net, a unified framework that integrates deep convolutional networks with variational autoencoders, utilizing a novel Feature-Adaptive Approximation Mechanism and variationally regularized few-shot inference. On CIFAR-100, VDLF-Net improves supervised classification over the reported VGG-16 and ResNet-50 Enhanced baselines (Table~\ref{tab:standard_results}); on Mini-ImageNet, it exceeds the evaluated Prototypical and Matching Networks under the same episodic protocol (Table~\ref{tab:fewshot_results}), with component ablations clarifying internal scale and loss contributions (Table~\ref{tab:ablation_min}). Broader claims about domains discussed in the introduction (e.g., medical imaging or satellite imagery) are not established by these benchmarks and remain directions for future empirical study.

\textbf{Future work.} (1)~Add modern few-shot baselines and multi-seed reports under Table~\ref{tab:hyperparams}; (2)~lighter gates or distillations for large-scale deployment; (3)~transfer the fusion stack to detection/segmentation or vision-language backbones.

\section*{Ethics Statement}
\label{sec:ethics}
This study uses only publicly available benchmarks (CIFAR-100 and Mini-ImageNet). No human participants were enrolled, and no proprietary clinical data were used. Ethics review for public benchmark use follows each provider's standard terms.

\section*{Conflict of Interest}
\label{sec:coi}
The authors declare no known competing financial interests or personal relationships that could bias the reported work.

\section*{Data Availability}
\label{sec:data_availability}
CIFAR-100 and Mini-ImageNet are publicly available. Splits follow Section~\ref{sec: Experimental Design}. Mini-ImageNet uses the usual 64/16/20 train/validation/test class partition.
Source code and configs to reproduce the experiments may be shared on reasonable request or with the camera-ready version, per the target journal's policy.

\bibliography{sn-bibliography}

@article{Litjens_2017,
  title   = {A Survey on Deep Learning in Medical Image Analysis},
  author  = {Litjens, Geert and Kooi, Thijs and Bejnordi, Babak Ehteshami and Setio, Arnaud Arindra Adiyoso and Ciompi, Francesco and Ghafoorian, Mohsen and van der Laak, Jeroen A. W. M. and van Ginneken, Bram and S{\'a}nchez, Clara I.},
  journal = {Medical Image Analysis},
  volume  = {42},
  pages   = {60--88},
  year    = {2017},
  doi     = {10.1016/j.media.2017.07.005}
}

@article{2021A,
  title   = {A Review of Deep Learning Methods for Semantic Segmentation of Remote Sensing Imagery},
  author  = {Yuan, Xiaohui and Shi, Jianfang and Gu, Lichuan},
  journal = {Expert Systems with Applications},
  volume  = {169},
  pages   = {114417},
  year    = {2021},
  doi     = {10.1016/j.eswa.2020.114417}
}

@article{2017Squeeze,
  title   = {Squeeze-and-Excitation Networks},
  author  = {Hu, Jie and Shen, Li and Albanie, Samuel and Sun, Gang and Wu, Enhua},
  journal = {IEEE Transactions on Pattern Analysis and Machine Intelligence},
  volume  = {42},
  number  = {8},
  pages   = {2011--2023},
  year    = {2020},
  doi     = {10.1109/TPAMI.2019.2913372}
}

@article{2020Generalizing,
  title   = {Generalizing from a Few Examples: A Survey on Few-Shot Learning},
  author  = {Wang, Yaqing and Yao, Quanming and Kwok, James T. and Ni, Lionel M.},
  journal = {ACM Computing Surveys},
  volume  = {53},
  number  = {3},
  pages   = {1--34},
  year    = {2020},
  doi     = {10.1145/3386252}
}

@article{2024A,
  title   = {A Systematic Review of Few-Shot Learning in Medical Imaging},
  author  = {Pachetti, Eva and Colantonio, Sara},
  journal = {Artificial Intelligence in Medicine},
  volume  = {156},
  pages   = {102949},
  year    = {2024},
  doi     = {10.1016/j.artmed.2024.102949}
}

@inproceedings{kendall2017uncertainties,
  title     = {What Uncertainties Do We Need in Bayesian Deep Learning for Computer Vision?},
  author    = {Kendall, Alex and Gal, Yarin},
  booktitle = {Advances in Neural Information Processing Systems},
  year      = {2017}
}

@article{2014Auto,
  title   = {Auto-Encoding Variational Bayes},
  author  = {Kingma, Diederik P. and Welling, Max},
  journal = {arXiv preprint arXiv:1312.6114},
  year    = {2013},
  doi     = {10.48550/arXiv.1312.6114}
}

@inproceedings{JiaBTG16,
  title     = {Dynamic Filter Networks},
  author    = {Jia, Xu and De Brabandere, Bert and Tuytelaars, Tinne and Van Gool, Luc},
  booktitle = {Advances in Neural Information Processing Systems},
  pages     = {667--675},
  year      = {2016}
}

@inproceedings{snell2017prototypicalnetworksfewshotlearning,
  title     = {Prototypical Networks for Few-shot Learning},
  author    = {Snell, Jake and Swersky, Kevin and Zemel, Richard S.},
  booktitle = {Advances in Neural Information Processing Systems},
  year      = {2017}
}

@inproceedings{finn2017modelagnosticmetalearningfastadaptation,
  title     = {Model-Agnostic Meta-Learning for Fast Adaptation of Deep Networks},
  author    = {Finn, Chelsea and Abbeel, Pieter and Levine, Sergey},
  booktitle = {Proceedings of the 34th International Conference on Machine Learning},
  series    = {Proceedings of Machine Learning Research},
  volume    = {70},
  pages     = {1126--1135},
  year      = {2017},
  publisher = {PMLR},
  address   = {Sydney, Australia}
}

@article{Schmidhuber_2015,
  title   = {Deep Learning in Neural Networks: An Overview},
  author  = {Schmidhuber, J{\"u}rgen},
  journal = {Neural Networks},
  volume  = {61},
  pages   = {85--117},
  year    = {2015},
  doi     = {10.1016/j.neunet.2014.09.003}
}

@inproceedings{CiresanMMGS11,
  title     = {Flexible, High Performance Convolutional Neural Networks for Image Classification},
  author    = {Ciresan, Dan Claudiu and Meier, Ueli and Masci, Jonathan and Gambardella, Luca Maria and Schmidhuber, J{\"u}rgen},
  booktitle = {Proceedings of the Twenty-Second International Joint Conference on Artificial Intelligence},
  pages     = {1237--1242},
  year      = {2011},
  doi       = {10.5591/978-1-57735-516-8/IJCAI11-210}
}

@inproceedings{HeZRS16,
  title     = {Deep Residual Learning for Image Recognition},
  author    = {He, Kaiming and Zhang, Xiangyu and Ren, Shaoqing and Sun, Jian},
  booktitle = {Proceedings of the IEEE Conference on Computer Vision and Pattern Recognition},
  pages     = {770--778},
  year      = {2016},
  doi       = {10.1109/CVPR.2016.90}
}

@inproceedings{vinyals2016matching,
  title     = {Matching Networks for One Shot Learning},
  author    = {Vinyals, Oriol and Blundell, Charles and Lillicrap, Timothy and Kavukcuoglu, Koray and Wierstra, Daan},
  booktitle = {Advances in Neural Information Processing Systems},
  year      = {2016}
}

@inproceedings{sung2018learning,
  title     = {Learning to Compare: Relation Network for Few-Shot Learning},
  author    = {Sung, Flood and Yang, Yongxin and Zhang, Li and Xiang, Tao and Torr, Philip H. S. and Hospedales, Timothy M.},
  booktitle = {Proceedings of the IEEE Conference on Computer Vision and Pattern Recognition},
  pages     = {1199--1208},
  year      = {2018},
  doi       = {10.1109/CVPR.2018.00131}
}

@inproceedings{chen2019closerlook,
  title     = {A Closer Look at Few-Shot Classification},
  author    = {Chen, Wei-Yu and Liu, Yen-Cheng and Kira, Zsolt and Wang, Yu-Chiang Frank and Huang, Jia-Bin},
  booktitle = {International Conference on Learning Representations},
  year      = {2019}
}

@inproceedings{triantafillou2020meta,
  title     = {{Meta-Dataset}: A Dataset of Datasets for Learning to Learn from Few Examples},
  author    = {Triantafillou, Eleni and Zhu, Tyler and Dumoulin, Vincent and Lamblin, Pascal and Evci, Utku and Xu, Kelvin and Goroshin, Ross and Gelada, Carles and Swersky, Kevin and Manzagol, Pierre-Antoine and Larochelle, Hugo},
  booktitle = {International Conference on Learning Representations},
  year      = {2020}
}

\end{document}